\documentclass{article}
\usepackage{spconf,amsmath,graphicx}
\usepackage{microtype}
\usepackage{multirow, multicol}
\usepackage{cite}

\title{Weakly supervised spoken term discovery \\using cross-lingual side information}
\makeatletter
\def\name#1{\gdef\@name{#1\\}}
\makeatother \name{{\em Sameer Bansal$^{1}$, Herman Kamper$^{1,2}$, Sharon Goldwater$^1$, Adam Lopez$^{1}$}}
\address{$^1$Institute for Language, Cognition, and Computation\\ $^2$Centre for Speech Technology Research\\School of Informatics, University of Edinburgh, UK \\
         {\tt\small \{s.bansal-2, h.kamper\}@sms.ed.ac.uk, \{sgwater, alopez\}@inf.ed.ac.uk}}

\ninept

\begin{document}
%
\maketitle
\begin{abstract}
Recent work on {\em unsupervised term discovery} (UTD) aims to identify and cluster repeated word-like units from audio alone. These systems are promising for some very low-resource languages where transcribed audio is unavailable, or where no written form of the language exists. However, in some cases it may still be feasible (e.g., through crowdsourcing) to obtain (possibly noisy) {\em text translations} of the audio. If so, this information could be used as a source of side information to improve UTD. Here, we present a simple method for rescoring the output of a UTD system using text translations, and test it on a corpus of Spanish audio with English translations. We show that it greatly improves the average precision of the results over a wide range of system configurations and data preprocessing methods.

\end{abstract}
\section{Introduction}
\label{sec:intro}
High-quality automatic speech recognition (ASR) systems require hundreds of hours of transcribed training data. As a result, they are currently available for only a tiny fraction of the world's several thousand languages~\cite{besacier2014}. To broaden accessibility, research on {\em zero-resource} speech technology aims to develop useful systems, such as unsupervised term discovery~\cite{park2008unsupervised,zhang+glass_icassp10,jansen2011efficient} or query-by-example~\cite{zhang+etal_icassp12,metze+etal_icassp13,levin2015segmental}, without the need for transcribed audio data. While considerable progress has been made in this area recently~\cite{versteegh+etal_interspeech15,rasanen+etal_interspeech15,kamper2015fully,kamper+etal_taslp16}, learning from audio alone is very challenging. Here, we ask whether using side information could improve performance.

In particular, we address the task of {\em unsupervised term discovery} (UTD), which aims to identify and cluster repeated word-like units from audio. We show that UTD can be improved using side information from text translations of the audio into another language. Such translations 
can often be obtained rapidly through crowd-sourcing, for example in disaster relief scenarios such as the 2010 Haiti earthquake~\cite{munro2010}. And when the low-resource language has no written form, text translations (ideally into a related language, as in~\cite{AlanBlackStuff}) may be considerably easier to obtain than a phonetic transcription.

In addition to improving UTD, our work may feed into the development of
cross-lingual tools for low-resource languages, in particular systems for {\em translating} speech from a low-resource language to a higher-resource language. A traditional pipeline would use ASR to transcribe the audio, followed by machine translation of the transcriptions. However, training such a system requires both transcribed audio in the low-resource language and parallel text.  Recent work~\cite{duong2015attentional} 
has begun to explore how to 
translate key words and phrases based on the kind of training data we use here. Our work could inform future approaches to this task.

Although our ultimate goal is to work with truly low-resource languages, ours is the first attempt we know of to address this task setting, so as a proof of concept we present results using a dataset of Spanish speech paired with English text translations~\cite{post2013improved}.
We use an open-source UTD system to discover potential word-like units from the audio, 
then use a simple rescoring method to improve the UTD output based on the translation information.
Our results
show large improvements in average precision across a wide range of hyperparameter settings, and also across cross-speaker matches.


\section{Unsupervised Term Discovery} \label{sec:utd}
Zero-resource speech technology addresses a number of different problems, ranging from automatic discovery of subword units~\cite{lee+glass_acl12,siu+etal_csl14} and improved feature representations~\cite{badino+etal_interspeech15,thiolliere+etal_interspeech15} to full segmentation and clustering of the audio into word-like units~\cite{walter+etal_asru13,lee+etal_tacl15,rasanen+etal_interspeech15,kamper+etal_taslp16}. Unsupervised term discovery is one of the most well-developed areas. 
Essentially, UTD systems search for pairs of audio segments that are similar, as measured by their dynamic time warping (DTW)~\cite{sakoe1978dynamic} distance. This task is inherently quadratic in the input size, and early systems~\cite{park2008unsupervised,zhang2009unsupervised} were prohibitively slow. Here, we use the open-source implementation in the Zero Resource Toolkit (ZRTools)\footnote{\texttt{https://github.com/arenjansen/ZRTools}}~\cite{jansen2011efficient}, a state-of-the-art system which uses a more efficient two-pass approach. It is also the only freely available UTD system we know of.

\subsection{Overview of the ZRTools UTD system}

\begin{figure*}
    \centering
    \centerline{\includegraphics[width=0.9\linewidth]{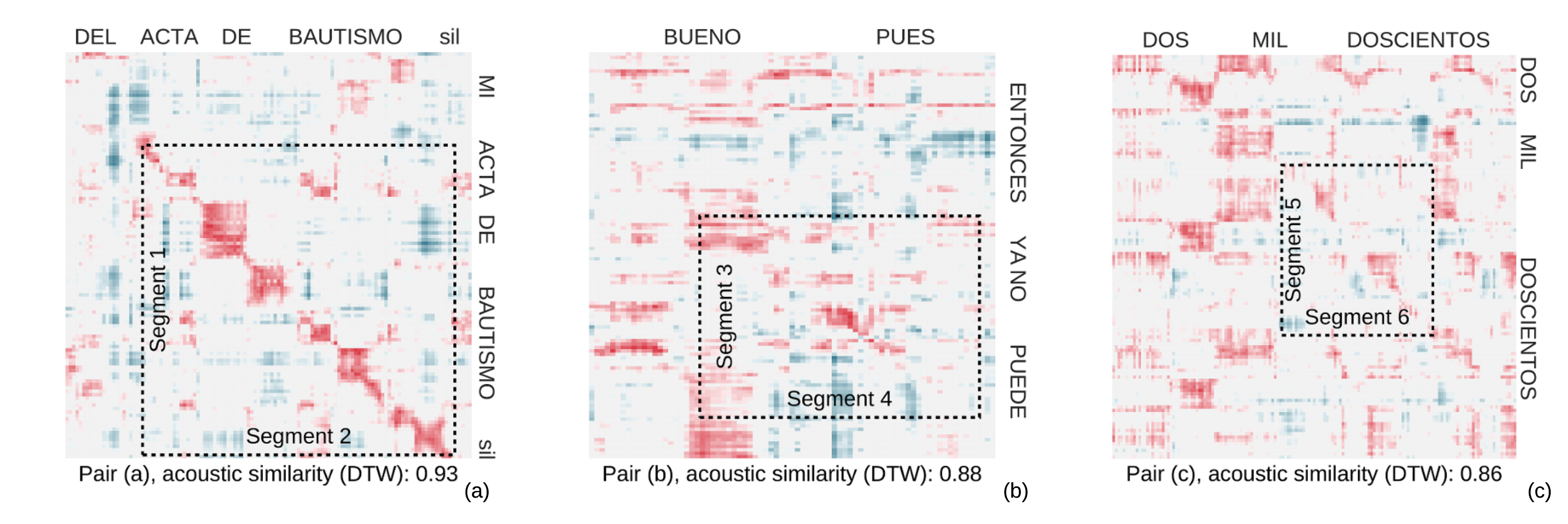}}
    \vspace*{-0.3\baselineskip}
    \caption{Acoustic similarity for utterance pairs. Dark/Red regions indicate strong match, Light/Blue indicate weak match. Dotted boxes mark the matching segments returned by UTD.
    (a) True Positive: Strong score for correct match. (b) False Positive: Strong score for incorrect match. (c) False Negative: Weak score for correct match. (Utterances are cropped for space so not all words are shown.)}
    \label{fig:dtws}
    \vspace*{-0.7\baselineskip}
\end{figure*}

In its first pass, ZRTools uses an approximate randomized algorithm and image processing techniques to extract potential matching segments. Image processing is used based on the intuition that if we plot the cosine similarity between every frame of the input feature vector representation (e.g. MFCCs), any repeated segments in the pair of utterances will show up as diagonal line patterns. Figure~\ref{fig:dtws}(a) illustrates this, showing a clear diagonal pattern corresponding to similar words in two utterances. 

In its second pass, ZRTools computes a normalized DTW score over potential matches to extract the final output. It returns segment pairs longer than a minimum duration (we used the recommended value of 500ms) along with their DTW score (between 0 and 1, with higher scores indicating greater similarity). These word-like or phrase-like segments can then be used for downstream tasks like keyword search and topic modeling~\cite{dredze2010nlp, zhang2009unsupervised, park2008unsupervised}.

Full details of the system can be found in~\cite{jansen2011efficient}.


\subsection{Limitations of UTD} \label{sec:zrterrors}
Like all UTD systems, ZRTools identifies patterns using acoustic information only. This can lead to various types of errors which we hope to reduce using cross-lingual side information.

\subsubsection{Mismatch between acoustic and semantic information} 
\label{subsec:phoneconfusion}
Some phonetically similar pairs identified by UTD are nevertheless different semantically, as illustrated in Figure~\ref{fig:dtws}(b). 
The words in the utterances are different, but UTD identifies similar phoneme sequences \texttt{a n o p w e} and \texttt{e n o p w e}.
Acoustic information alone cannot overcome these errors, yet they will cause semantic errors in downstream applications such as machine translation or spoken document retrieval.

On the other hand, due to noise and variability (both within and across speakers), not all semantically correct matches will be assigned a high DTW score. The ZRTools documentation recommends using a DTW score threshold of 0.88 to filter good matches, yet there are many correct pairs with scores lower than 0.88. One example is illustrated in Figure~\ref{fig:dtws}(c), where a correct match has a score just below the cut-off threshold. Of course, we can lower the DTW threshold to return more pairs (raise recall), but this will also increase the number of incorrect pairs returned (lower precision).

\subsubsection{Silence and filler words as valid matches} \label{subsec:pausewords}
UTD is sensitive to silence regions, background noises, and filler words, 
all three of which
commonly occur in conversational speech. We observed that these phenomena generate a large number of discovered pairs, since they are frequent and are often good acoustic matches with each other. 

The number of non-word pairs found by UTD due to these phenomena depends on the preprocessing of the data. To show that our method improves UTD output regardless of preprocessing, we experiment with two different preprocessing methods.

First, we use the automatic voice activity detection (VAD) script that comes with ZRTools, which uses Root Mean Squared (RMS) Energy detection to label VAD regions. Only these regions are then used when searching for patterns. 
This method aggressively filters out silence but
removes a considerable amount of valid speech. It also retains many filler words, which often have high energy and long duration.


Alternatively, we can use a forced alignment (FA) of the speech data with the transcripts to filter out non-speech regions. This method would not be available in a true zero-resource situation, but might better reflect the output of a more sophisticated automatic VAD system, and has the advantage of retaining more of the training data. However, we observed that it removes fewer silent regions than the VAD system.

Regardless of which preprocessing method is used, UTD will tend to find a large number of matches based on non-word regions. However, the translations for such non-word pairs will rarely contain any content words in common, so rescoring them based on their translations should reduce these spurious matches.

\section{Improving UTD using translations}
\label{sec:utdwithtranslations}

Given pair of speech utterances, we hypothesize that the \textbf{similarity in their translations provides a (noisy) signal of the semantic similarity between the discovered acoustic units, and that this signal can improve UTD.}

Consider the examples in Table~\ref{tab:comparison}, which shows the English translations of the utterances containing the segments depicted in Figure~\ref{fig:dtws}. (The utterances are cropped, so not all Spanish words are shown.)
Stop words are shown in parentheses; we filter these out using the NLTK toolkit\footnote{\texttt{http://www.nltk.org/}} before computing translation similarity.
Notice that the two pairs (a and c) that have matching Spanish words also have matching English content words, even though one of them falls below the recommended 0.88 threshold for a UTD (acoustic) match. On the other hand, pair (b) has a high UTD score due to phonetic similarity, but there is no match between the English words.




\begin{table}
\centering
\begin{tabular}{@{\extracolsep{-4pt}} llp{4cm}ll }
 Pair & Seg & English translation & \parbox{1cm}{Acou. sim.} & \parbox{1cm}{Transl. sim.} \vspace{3pt}\\
 \hline
  \multirow{2}{1em}{a} & 1 & (to) tell (them) (to) send (me) (my) \textbf{baptism} act  & \multirow{2}{2em}{0.93} & \multirow{2}{2em}{0.125}\\ 
 & 2 & (we) (are) going (to) need (the) sacrament (of) \textbf{baptism} paper & & \\
 \hline
 \multirow{2}{1em}{b} & 3 & (not) (now) (now) (then) (he) cant anymore & \multirow{2}{2em}{0.88} & \multirow{2}{2em}{0} \\
 & 4 & yes well (its) good well yeah & & \\
 \hline
 \multirow{2}{1em}{c} & 5 & okay (this) (the) address \textbf{two thousand two hundred} & \multirow{2}{2em}{0.86} & \multirow{2}{2em}{0.600}\\
 & 6 & \textbf{two thousand two hundred} & & \\
 \hline
\end{tabular}
\caption{English translations for the utterance pairs shown in Figure~\ref{fig:dtws}. The acoustic (DTW) similarity and translation (Jaccard) similarity for each pair is also shown.  Stop words (in parentheses) are not used to compute translation similarity. Matching content words in \textbf{bold}.}
\label{tab:comparison}

\end{table}


To exploit these observations, we rescore the pairs returned by ZRTools using their translation similarity.
If $dtw_i$ is the acoustic similarity score for pair $i$ computed by ZRTools, and $J_i$ is the translation similarity score (described below), then the new score of pair $i$ is computed as the $\alpha$-weighted mean between the two:
\begin{align} \label{eq:scoreupdate}
score_i &= (1-\alpha)~\times~dtw_i~+~\alpha~\times~J_i
\end{align}

To compute the similarity $J$ between a pair of English translations, we treat each translation as bag of words (after filtering for stop words), and use Jaccard similarity~\cite{jaccard1901distribution}:
\begin{align} \label{eq:jaccardsim}
    J &= \frac{| E_1~\cap~E_2 |}{| E_1~\cup~E_2 |}
\end{align}
where $E_1$ is the set of content words in translation 1 and $E_2$ is the set of content words in translation 2.


Note that even seemingly low translation similarity scores (such as 0.125 for pair (a) in Table~\ref{tab:comparison}) are
still a strong signal of semantic similarity between acoustic matches, because any non-zero score indicates some content words in common. Empirically we have observed $J~\ge~0.1$ to be a good indicator of a correct match (although in practice we do not impose any threshold on $J$).


\section{Experimental setup and evaluation}
\label{sec:expsetup}

In all experiments, the input consists of speech from the CALLHOME Spanish corpus and the crowdsourced English translations of~\cite{post2013improved}. The corpus consists of speech from telephone conversations between native speakers. We use the default feature representation as used by ZRtools: 39-dimensional Relative Spectral Transform - Perceptual Linear Prediction (PLP) feature vectors. 

We carry out four sets of experiments as summarized in Table~\ref{tab:expconfigs}.
Note that the energy-based VAD filters out far more of the data than forced alignment.
For each phone call, we have two channels of audio, each with at least one speaker, but sometimes more. The same speaker may be on multiple calls. However, for the purposes of our cross-speaker evaluation, we assume that each channel corresponds to a unique speaker.

\begin{table}[t]
\centering
\begin{tabular}{ lll }
 \# conversations & Preprocessing & Active speech (hrs)\\ 
 \hline
 20 & VAD & 0.91 \\ 
 50 & VAD & 2.29 \\ 
 20 & forced alignments (FA) & 2.80 \\
 50 & forced alignments (FA) & 6.92 \\
\end{tabular}
\caption{The four sets of input speech data used for UTD showing total speech after preprocessing by VAD or forced alignment.
}
\label{tab:expconfigs}
\end{table}

To evaluate the results of the raw UTD system and our rescoring method, we
use the original Spanish CALLHOME transcripts to check if a pair of discovered speech segments is actually a true match. Note that the transcripts are not otherwise used as input to our system, except to filter non-speech in the forced alignment setting, where it serves as a kind of oracle for speech detection. For each pair of segments, we retrieve the corresponding words (as per the time stamps) from the transcripts.
We retrieve any words which either partially or completely overlap with ZRTools output. The retrieved words are then filtered for stop words using NLTK. A discovered pair is marked as correct if the two segments have at least one content word in common; otherwise, it is marked as incorrect.


To implement our rescoring method, we begin by running UTD with an acoustic matching threshold of $D=0.8$, which is considerably lower than the ZRTools recommended level of $D=0.88$. 
An empirical check suggested that very few correct pairs had scores below 0.8, and this value of $D$ gives us enough potential pairs to perform rescoring.

For evaluation, we treat the set of correct pairs returned with $D=0.8$ as the total number of possible correct pairs---that is, recall values are computed with respect to this number. Therefore, a recall value of 1 does {\em not} mean that all correct pairs in the entire dataset have been identified, only those whose DTW score is above 0.8.

Using our recomputed scores (as defined in Equation~\ref{eq:scoreupdate}) we can choose a new threshold value $S$, and return pairs that score above $S$. For each value of $S$, we can compute precision and recall:
\begin{align} \label{eq:precision}
	\mbox{Precision@$S$} &= \frac{\sum_{i=1}^{N} (correct_i \land score_i~\ge~S)}{\sum_{i=1}^{N} (score_i\ge S)}
\end{align}
\begin{align} \label{eq:recall}
	\mbox{Recall@$S$} &= \frac{\sum_{i=1}^{N} (correct_i \land score_i~\ge~S)}{\sum_{i=1}^{N} (correct_i \land dtw_i~\ge~0.80)}
\end{align}
where $correct_i$ indicates if a UTD output pair is correct or not, and $N$ is the total number of pairs discovered with DTW threshold $D=0.80$.
We find the Precision/Recall curve by considering all possible values of $S$, and then 
compute the \textbf{average precision} (AP) as the area under this curve.

\section{Results and discussion}
\label{sec:discussion}

\begin{table}[t]
  \centering
\begin{tabular}{@{\extracolsep{6pt}} l@{}r@{\,/\,\hspace{-6pt}}l@{}rr@{\,/\,\hspace{-6pt}}l@{}r }
  & \multicolumn{3}{c}{$D=0.80$} & \multicolumn{3}{c}{$D=0.88$} \\
  \cline{2-4} \cline{5-7}
 Config & Corr. &Tot. & xspk & Corr. & Tot. & xspk \\
 \hline
 20, VAD & 66 & 206 \,(.32) & 5 & 52 & 90 \,(.58) & 1 \\
 50, VAD & 182 & 1330 \,(.14) & 35 & 138 & 490 \,(.28) & 11 \\
 20, FA & 1064 & 7541 \,(.14) & 114 & 728 & 2339 \,(.31) & 18 \\
 50, FA & 3119 & 43762 \,(.07) & 963 & 1918 & 11114 \,(.17) & 287 \\

\end{tabular}
\caption{Number of pairs returned by ZRTools for different thresholds ($D=0.8, D=0.88)$ and input configurations (20 or 50 conversations, with non-speech regions filtered using energy-based VAD or forced alignments). \textbf{Corr./Tot.} is the number of correct and total pairs discovered, yielding precision values in parentheses. \textbf{xspk} is the number of correct pairs discovered across speakers.}
\label{tab:utdout}
\end{table}

\subsection{Baseline UTD system} \label{subsec:vadcomparison}

Table~\ref{tab:utdout} lists the number of pairs discovered by running the baseline UTD system in each configuration. The number of pairs discovered using energy-based VAD is low, as expected, due to large parts of speech data being filtered out. Using forced alignments gives us a higher number of discovered pairs, but
at a cost of precision.
The low number of cross-speaker pairs listed in Table~\ref{tab:utdout} highlights the difficulty of discovering these, though they are important for downstream tasks~\cite{kamper2015unsupervised,renshaw+etal_interspeech15}. Translation information may be particularly helpful for identifying cross-speaker pairs, but our method is limited by the small number of pairs that are discovered in the first place, as shown in the xspk columns in Table~\ref{tab:utdout}. In future work we plan to investigate whether translation information could be fed into the UTD system at an earlier stage to help discover more cross-speaker pairs.


\subsection{Improvements using translations} \label{subsec:results}

\begin{figure}[t]
    \centering
    \centerline{\includegraphics[width=0.95\linewidth]{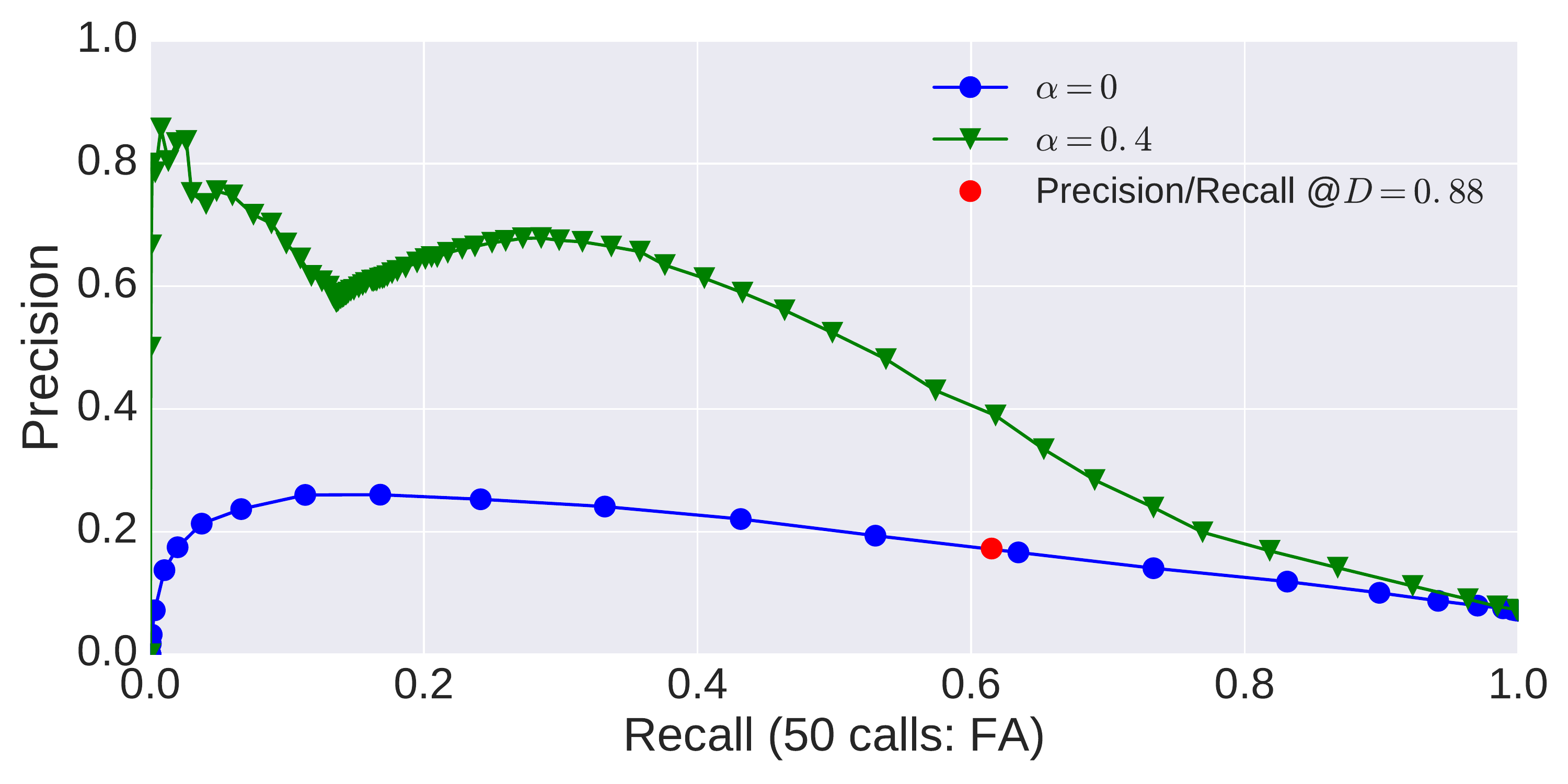}}
    \vspace*{-0.3\baselineskip}
    \caption{Precision/Recall curves for the \texttt{50, FA} configuration, with $\alpha=0$ (baseline UTD) and $\alpha=0.4$ (our system).  AP is computed as the area under the respective curves.}
    \label{fig:pr_50_align}
    \vspace*{-0.7\baselineskip}
\end{figure}

Figure~\ref{fig:pr_50_align} illustrates the benefit of our system, showing Precision/Recall curves for the (\texttt{50,~FA}) setting. By using only acoustic information and varying the value of $D$, only points on the lower (blue) curve can be achieved (e.g., the red point is for $D=0.88$). Using translations (here $\alpha=0.4$) clearly improves results.


\begin{figure}
    \centering
    \centerline{\includegraphics[width=0.95\linewidth]{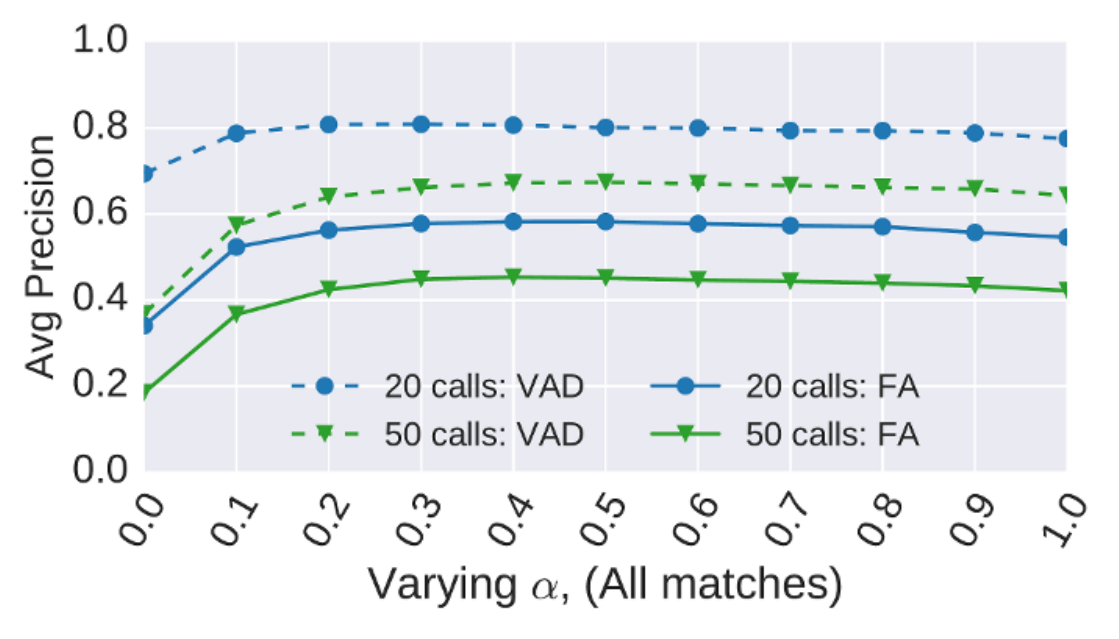}}
    \vspace*{-0.3\baselineskip}
    \caption{Average precision for various configurations, varying $\alpha$ from 0 (translations not used) to 1 (only using translation similarity).}
    \label{fig:results}
    \vspace*{-0.7\baselineskip}
\end{figure}

To show that these benefits are not highly sensitive to $\alpha$,
Figure~\ref{fig:results} plots the AP for all configurations listed in Table~\ref{tab:expconfigs}, for $\alpha$ between 0 and 1.
For every configuration and every $\alpha~>~0$ setting, we obtain higher AP than the baseline $\alpha~=~0$, often by a large margin.
AP scores for $\alpha =0.4$, which is one of a range of good values, are
 listed in Table~\ref{tab:metrics40}.
Note that the AP numbers are only comparable between systems using the same data/preprocessing configuration, since the total number of pairs that need to be discovered to achieve 100\% recall is different in each case (it is given by the number of correct pairs at $D=0.8$ in Table~\ref{tab:utdout}).

\begin{figure}[t]
    \centering
    \centerline{\includegraphics[width=0.95\linewidth]{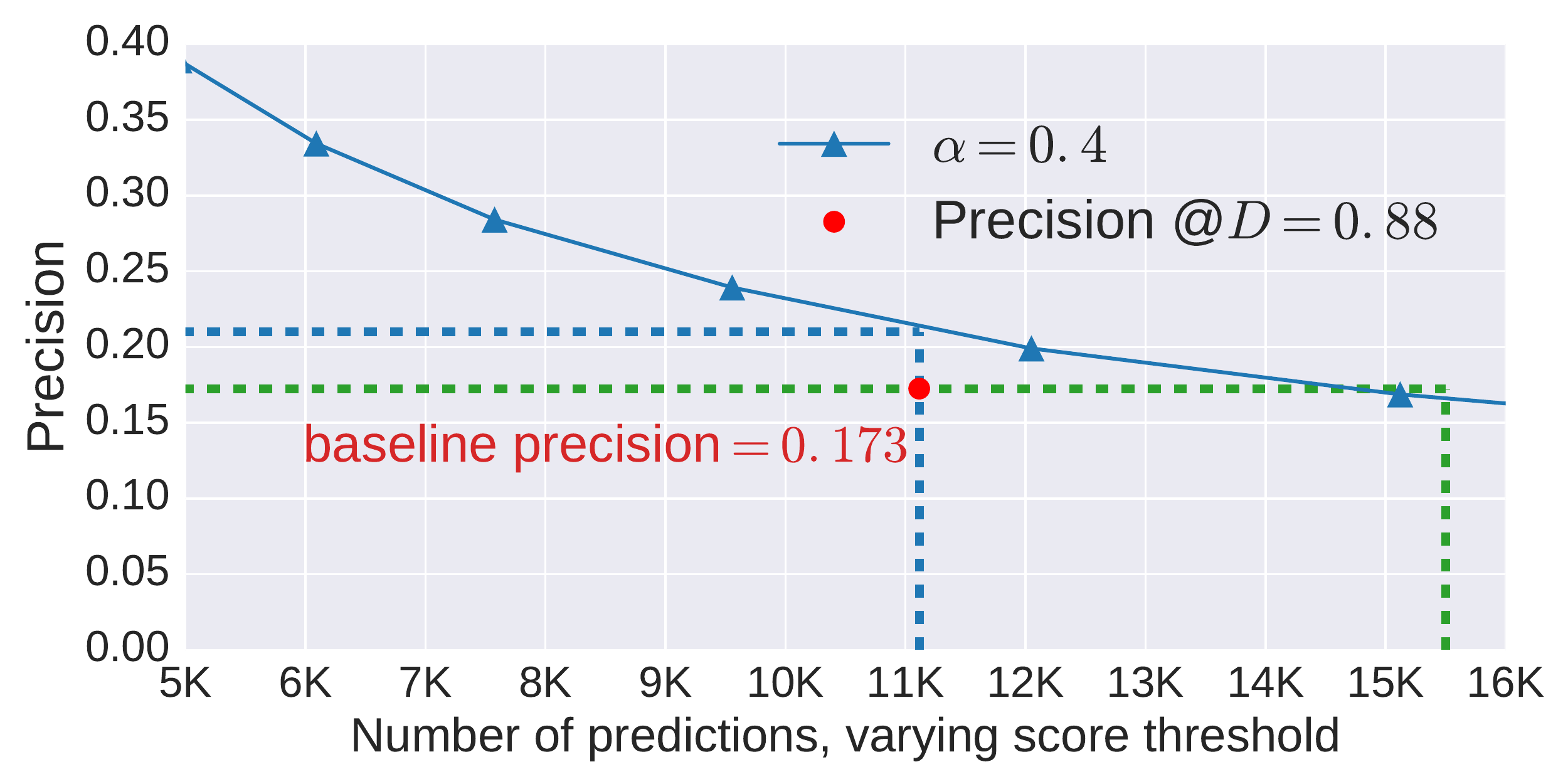}}
    \vspace*{-0.3\baselineskip}
    \caption{Predictions/Precision plot for $\alpha=.4$ (50 conversations, with non-speech regions filtered using forced alignments). The dot marker indicates the precision with recommended $D=0.88$. The plot shows that our system has better precision.}
    \label{fig:prec_pred}
    \vspace*{-0.7\baselineskip}
\end{figure}

Figure~\ref{fig:prec_pred} compares our system directly to the UTD system's recommended setting of $D=0.88$, which returns about 11K matches yielding a precision of $0.17$. If we set $S$ in our system to return the same number of matches, the precision rises to $0.21$. We similarly found that for cross-speaker matches, at 6.9K predictions (the baseline output), our system using translation improves precision from 0.04 to 0.07.

  

\begin{table}[t]
\centering
\begin{tabular}{lllll}
 \multirow{2}{3em}{Config} & \multicolumn{2}{l}{AP: all matches} & \multicolumn{2}{l}{AP: cross-spkr only} \\
 & $\alpha=0$ & $\alpha=0.4$ & $\alpha=0$ & $\alpha=0.4$ \\
 \hline
 20, VAD & 0.694 & 0.806 & 0.089 & 0.375 \\
 50, VAD & 0.368 & 0.672 & 0.066 & 0.322 \\
 20, FA & 0.341 & 0.583 & 0.025 & 0.130 \\
 50, FA & 0.185 & 0.454 & 0.036 & 0.128
\end{tabular}
\caption{\textbf{AP}: Average precision results for $\alpha=0$ (baseline UTD) and our system with $\alpha=0.4$. Including results for all matches, and Cross-Speaker only.}
\label{tab:metrics40}
\end{table}


Figure~\ref{fig:results} shows that at $\alpha=1$, the AP is higher than at $\alpha=0$, which seems to imply that ignoring the DTW score and only using translation similarity yields better results. However, recall that we started by pruning the ZRTools output using a DTW threshold of~$0.80$, so even with $\alpha=1$, our system is not actually ignoring acoustic information. In addition, the UTD system provides important information about segment boundaries which translations alone cannot.

As discussed in Section~\ref{subsec:pausewords}, we observed that UTD discovers many filler words with high DTW score, but their translation score should be low. Unfortunately it is difficult to quantify how well our system filters out these matches since filler words are usually not transcribed in our data.



\section{Conclusion}
\label{sec:conclusion}
We have shown that side information in the form of translations improves the output of UTD across a wide range of settings. In future work, we will use the improved UTD output to learn better cross-speaker speech features for low-resource settings, and explore the use of translations as a preprocessing step for UTD, by helping guide the search for matches. We also aim to expand this work into a semi-supervised setting, using additional unlabeled speech data to improve UTD. We believe this can be done using approaches such as label propagation~\cite{zhu2002learning} and label spreading~\cite{bengio2006label,zhou2004learning}.

\section{Acknowledgments}
\label{sec:thanks}
We thank David Chiang and Antonios Anastasopoulos for sharing alignments of the CALLHOME speech and transcripts; Aren Jansen for assistance with ZRTools; and Marco Damonte, Federico Fancellu, Sorcha Gilroy, Ida Szubert, and Clara Vania for comments on previous drafts. This work was supported in part by a James S McDonnell Foundation Scholar Award and a Google faculty research award.



\vfill\pagebreak
\bibliographystyle{IEEEbib}
\bibliography{icassp_2017.bib}

\end{document}